\documentclass[10pt,twocolumn,letterpaper]{article}

\usepackage{cvpr}
\usepackage{times}
\usepackage{epsfig}
\usepackage{graphicx}
\usepackage{amsmath}
\usepackage{amssymb}

\usepackage[dvipsnames,svgnames]{xcolor}
\usepackage{enumitem}
\usepackage{xspace} 
\usepackage{multirow}
\usepackage{bbm}
\usepackage{url}
\usepackage{lettrine}
\usepackage[ruled,vlined,linesnumbered]{algorithm2e}
\usepackage{color}

\usepackage[breaklinks=true,bookmarks=false]{hyperref}

\cvprfinalcopy 

\def\cvprPaperID{2780} 

\begin{document}

\title{
Unsupervised part learning for visual recognition
}

\author{Ronan Sicre$^1$, Yannis Avrithis$^1$, Ewa Kijak$^1$ and Fr\'ed\'eric Jurie$^2$\\
$^1$ INRIA / IRISA  
$^2$ Normandie Univ, UNICAEN, ENSICAEN, CNRS - UMR GREYC\\
{\tt\small \{ronan.sicre, ioannis.avrithis\}@inria.fr  ewa.kijak@irisa.fr  frederic.jurie@unicaen.fr}
}

\maketitle

\begin{abstract}
Part-based image classification aims at representing categories by small sets of learned discriminative parts, upon which an image representation is built. Considered as a promising avenue a decade ago, this direction has been neglected since the advent of deep neural networks. In this context, this paper brings two contributions: 
first, this work proceeds one step further compared to recent part-based models (PBM), focusing on how to learn parts without using any labeled data. Instead of learning a set of parts per class, as generally performed in the PBM  literature, the proposed approach both constructs a partition of a given set of images into visually similar groups, and subsequently learns a set of discriminative parts per group in a fully unsupervised fashion. 
This strategy opens the door to the use of PBM in new applications where labeled data are typically not available, such as instance-based image retrieval. 
Second, this paper shows that despite the recent success of end-to-end models, explicit part learning can still boost classification performance.
We experimentally show that our learned parts can help building efficient image representations, which outperform state-of-the art Deep Convolutional Neural Networks (DCNN) on both classification and retrieval tasks.

\end{abstract}

\newcommand{\head}[1]{{\smallskip\noindent\bf #1}}
\newcommand{\alert}[1]{{\color{red}{#1}}}
\newcommand{\old}[1]{{\color{blue}{#1}}}

\def\proj{\operatorname{Proj}}
\def\tr{\operatorname{tr}}
\def\vec{\operatorname{vec}}

\newcommand{\card}[1]{\left|{#1}\right|}
\newcommand{\dotp}[2] {\left\langle {#1}, {#2} \right\rangle}
\newcommand*{\lcdot}{\hspace{-0.1ex}\raisebox{-0.7ex}{\scalebox{2}{$\cdot$}}\hspace{-0.2ex}}
\newcommand{\norm}[1] {\|{#1}\|}

\newcommand{\real}{\mathbb{R}}
\newcommand{\one}{\mathbf{1}}
\newcommand{\ind}[1]{\mathbbm{1}_{#1}}

\newcommand{\cA}{\mathcal A}
\newcommand{\cB}{\mathcal B}
\newcommand{\cI}{\mathcal I}
\newcommand{\cP}{\mathcal P}
\newcommand{\cR}{\mathcal R}
\newcommand{\cU}{\mathcal U}

\def\etal{{\em et al}\@\xspace}
\def\ie{{\em i.e.}\@\xspace}
\def\eg{{\em e.g.}\@\xspace}
\def\cf{{\em cf}\@\xspace}
\def\etc{{\em etc.}\@\xspace}
\def\vs{{\em vs.}\@\xspace}
\def\wrt{w.r.t\@\xspace}

\def\ISA{ISA\@\xspace}
\def\HunA{HunA\@\xspace}

\section{Introduction}
\label{sec:intro}

Part-based models -- \ie the family of models considering categories, objects, \etc as sets of elements that are meaningful, discrete, and limited in number -- offer several interesting properties for the representation of images. 
First of all, as they rely on limited meaningful sets of image regions, they explicitly provide strong cues for discovering images’ structures, \ie they explicitly break images into useful components. 
In addition, they also provide more compact representations than methods based on the pooling of large numbers of regions, as the number of parts is generally low compared to the number of different regions an image contains. 
Finally, as they focus on the {\em key} parts of images, they are expected to give image representations better suited, in terms of performance, to computer vision tasks such as image classification, recognition, or retrieval. 

Owing to their attractive properties,
part-based models were addressed extensively in the past, some of the major representatives being the {\em constellation model} \cite{weber10}, Ullman's {\em fragment-based model} \cite{Ullman2001gh},  or the  {\em Interleaved Categorization and Segmentation model} \cite{Leibe2008}.

Despite the relative success of these works, it has to be recognized that the recent success of deep convolutional neural networks (DCNN)  raised a tsunami which swept away most of the past models, leaving space only for statistical models that use very dense sampling of image regions, and alternate between pooling and convolutional steps. 
Such models, incredibly good in terms of performance, have heavy computational costs and require massive amounts of labeled data.

In this context, one contribution of this paper is to address one strong limitation of most of the existing part-based models, namely the necessity to rely on annotated images to learn (or discover) task specific parts. This supervised part learning stage  is crucial in most of the past methods and prevent their use in tasks for which labeled data are not available, \eg image retrieval.

Another contribution of this paper is to bring a strong empirical evidence that part-based models can exceed the performance of DCNNs representations for both classification and image retrieval tasks.
More precisely, we experimentally show that a part-based model can compete with  state-of-the-art DCNNs, encoding very dense representations of images.



This article demonstrates experimentally, on two classification tasks (Willow and MIT67) and on two retrieval tasks (Oxford5k and Paris6k ), that the proposed part-based representations, learned without any annotated images, can  efficiently encode the images, improving the performance of state-of-the art DCNN representations.

The rest of the paper is organized as follows: Section~\ref{sec:previous} introduces the related works, Section~~\ref{sec:method} exposes the proposed method while Section~\ref{sec:exp} presents the experimental validation.
\section{Previous work}
\label{sec:previous}

This section focuses on part-based models, and, more particularly on how these models learn parts. We can distinguish between the approaches making use of annotated data in the learning process and those which does not. We will discuss these two categories in turn. 

On the side of the approaches not using labeled images for learning the parts, the only work we are aware of is  the work of Singh \etal \cite{singh12}. In \cite{singh12}, parts are defined as  sets of relevant  patches that are frequent enough in addition to being discriminative w.r.t. other parts. The problem is formulated as an unsupervised discriminative clustering problem on a huge dataset of image patches, optimized by an iterative procedure alternating between clustering and training discriminative classifiers. 
Despite the interest of the method, the performance is, by far, not as good as supervised approaches described in the rest of the section.

Most of the past approaches define parts as image regions, allowing to discriminate efficiently between the different categories involved in the task. However, they differ in the way they select the candidate regions and in how they evaluate their ability to distinguish the categories. Ullman's fragment-based model \cite{Ullman2001gh} randomly samples candidate regions, detects parts by template matching and defines parts as templates, which are likely to be found in images of one class but not in images of the other ones (likelihood ratio).  In the constellation model, of \cite{weber10}, the variability within a class is represented by a joint probability density function on the shape of the constellation and the appearance of the parts. Distinctive features in the training set are learned with other model parameters using expectation maximization. It is assumed that only one category of image is present during training. In the Deformable Part Model, proposed by Felzenszwalb \etal \cite{felzenszwalb10}, the aforementioned questions are addressed  by selecting discriminative regions that have significant overlap with a given bounding box location. The association between regions and part is done through the estimation of some latent variables, \ie, the positions of the regions w.r.t. the position of the root part of the model.  

Doersch \etal \cite{doersch13} used density based mean-shift algorithms to discover discriminative regions. Starting from a weakly labeled image collection, coherent patch clusters that are maximally discriminative with respect to the labels are produced,  requiring a single pass through the data. More recently,  Juneja \etal \cite{Juneja13} also aimed at discovering distinctive parts for an object or scene class by first identifying the likely discriminative regions by low-level segmentation cues, and then learning part classifiers on top of these regions. The two steps are alternated iteratively  until a convergence criterion based on Entropy-Rank is satisfied. 
Similarly Mettes \etal \cite{Mettes15} propose to learn parts that are shared across classes.
The more recent approach of Sicre \etal \cite{Sicre14parts,Sicre15A} proposes to learn the parts during a softassign-like matching algorithm, building part representations as well as matching parts to regions from labelled images.  This work gives state of the art results on several datasets.

Beside these aforementioned approaches, which separate the classification process in two stages, one for learning the part and a second for learning the classifiers once the images are encoded, the recent approaches of \cite{Kulkarni16,parizi2014automatic} rely on a joint learning of all the parts and the category classifiers together. This joint learning approach of all components of part-based models is  particularly relevant since the discriminative regions are explicitly optimized for the targeted task. 

Approaches such as \cite{Kulkarni16,parizi2014automatic}, despite their excellent performance, put more weight on the need of annotated images to learn the parts, and also produce parts that are more strongly related to the categories. 
In this paper we aim at learning parts independently of categories, so they can be used for tasks for which no categories are defined (\eg, image retrieval), while giving comparable level of performance as jointly learned parts. 

In the context of image retrieval, part learning takes the form of an offline processing stage where patterns are automatically mined in images. In this sense, relevant works are the discovery of spatially related images~\cite{QuLV08} and their parts~\cite{ChMa10,HGO+10}, the discovery of favorite views of popular images~\cite{WeLe11}, the selection of local features based on pairwise matching~\cite{TuLo09},
the online aggregation of multiple query descriptors~\cite{Sicre15B},
or the offline aggregation of different views in scene representations~\cite{AKTS10}. Such methods may be used to improve image retrieval, even with state of the art CNN representations~\cite{Tolias16,RaTC16,gordo16}. In contrast to such works, we do not rely on pairwise matching or precise geometry verification, but we rather learn a joint representation of parts that are matched across images. Moreover, the parts are discriminative among different image groups.
\section{Method}
\label{sec:method}

The proposed approach builds on the recent work of~\cite{Sicre15A}, from which we borrow the idea of considering part discovery as an assignment problem, where assignment is between regions and parts. In contrast with \cite{Kulkarni16,parizi2014automatic}, decoupling part learning from the main task (\eg image classification) makes possible the learning of parts from raw images, independently of any category definition. 
As a high-level interpretation, the learned parts can be seen a vocabulary of latent discriminative mid-level features, which are later detected in images to generate image descriptions.

\subsection{Problem formulation}
\label{sec:formulation}

\head{Notation.}
%
%
Given matrices $A, B$ of the same size, $\dotp{A}{B} = \sum_{i,j} a_{ij} b_{ij}$ is their (Frobenius) inner product.
Vector $\one_n$ is an $n \times 1$ vector of ones.
Finally, $[n]$ is the set $\{1, \dots, n\}$.

\head{Images and groups.} Following~\cite{Sicre15A}, we denote by $\cI$ the set of training images, with $N=\card{\cI}$. Images in $\cI$ are denoted $I_n$ for $n \in [N]$. Unlike~\cite{Sicre15A}, we assume no category labels during part learning. Instead, the images $\cI$ are partitioned into $K$ groups,
\begin{equation}
	\cI=\bigcup_{k \in [K]}\cI^k,
	\label{eq:partition}
\end{equation}
%
where $\cI^k$ are the images of group $k$, with $N^k=\card{\cI^k}$. The partition $\cB = \{\cI^k: k \in [K]\}$ is unknown.

\head{Regions.}
A set of regions ${\cR}_I$ is extracted from each image $I \in \cI$. The number of regions per image is fixed and denoted $\card{\cR}$. The set of regions from images in group $k$ is denoted $\cR^k$, with $R^k = \card{\cR^k} = N^k\card{\cR}$ regions. The total number of regions in the training set $\cI$ is 
$R = N\card{\cR}$.
Given a matrix $A$ with columns indexed by regions, we denote by $A_I$ the submatrix that contains columns $r \in \cR_I$ corresponding to image $I$.

\head{Descriptors.}
Each region $r \in \cR_I$ is represented by a descriptor $x_r \in \real^d$, which is the output of a DCNN inner layer on the region $r$, see Section \ref{sec:details}. By $X$ ($X^k$) we denote the $d\times R$ ($d\times R^k$) matrix whose columns are the descriptors of all training images $\cI$ (group of images $\cI^k$).

\head{Parts.}
For each group of images $\cI^k$, we learn a set of parts $\cP^k$. We assume there is a fixed number $P = \card{\cP^k}$ of parts per group. Following~\cite{Sicre15A}, we use the $P \times R^k$ matrix $A^k$ associating image regions $\cR^k$ to parts. 
Ideally, element $a_{pr}^k=1$ if region $r$ represents part $p$, and $0$ otherwise.

\head{Requirements.}
We adjust the requirements of~\cite{Sicre15A} to an unsupervised setting: (i) in each group, the $P$ parts are different from one another, (ii) each part of $\cP^k$ is present in every image of its group $\cI^k$, (iii) parts in $\cP^k$ should occur more frequently in images in $\cI^k$ than in the remaining training images $\cI \setminus \cI^k$. The first two requirements define the following constraints on matrix $A^k$ for each group $k$:
\begin{align}
	\one_P^\top A^k         & \le \one_{R^k}^\top \label{eq:cons-different} \\
	A_I^k \one_{\card{\cR}} & = \one_P \,\, \text{ for } I \in \cI^k \label{eq:cons-present}
\end{align}
where $\le$ is meant element-wise.
This implies that each sub-matrix $A_I^k$ is a \emph{partial assignment} matrix. Then, the admissible set $\cA^k$ of matrices $A^k$ is the non-convex subset of $\{0,1\}^{P\times R^k}$ satisfying constraints~(\ref{eq:cons-different}) and (\ref{eq:cons-present}).

\head{Part models.}
The third requirement is modeled by Linear Discriminant Analysis (LDA): given matrix $A^k$ in group $k$, the model $w_p(A^k)$ of part $p \in \cP^k$ is defined as the $d$-vector
\begin{align}
	w_p(A^k) & \triangleq \Sigma^{-1}\left(
		\frac{\sum_{r\in\cR^k} a_{pr}^k x_r^k }{\sum_{r\in\cR^k} a_{pr}^k} - \mu
	\right) \label{eq:lda-scalar}
	,
\end{align}
where $\mu = \frac{1}{N} X\one_R$ and $\Sigma = \frac{1}{N} (X- \mu\one_R^\top)(X- \mu\one_R^\top)^\top$ are the empirical mean and covariance matrix of region descriptors over all training images. The classification score of a given region descriptor $x_r$ for the model of part $p \in \cP^k$ is then given as the inner product $\dotp{w_p(A^k)}{x_r}$.

The models of all parts $p\in \cP^k$ are concisely represented by $d\times P$ matrix
\begin{equation}
	W(A^k) \triangleq \Sigma^{-1} \left( \frac{1}{N^k} X^k (A^k)^\top - \mu \one_P^\top \right).
	\label{eq:lda-mat}
\end{equation}
whose columns are vectors $w_p(A^k)$ for $p \in \cP^k$. Then, the scores of all region descriptors $X^k$ for all parts in $\cP^k$ are given by the $P \times R^k$ matching matrix
\begin{equation}
	M(A^k) \triangleq W(A^k)^\top X^k.
	\label{eq:matching}
\end{equation}
\head{Objective function.} Given that part models are expressed as a function of matrix $A^k$ for each group $k$~(\ref{eq:lda-mat}), we are looking for an optimal matrix in the admissible set $\cA^k$,
\begin{align}
	(A^k)^\star & \in \arg\max_{A^k \in \cA^k} J(A^k) \label{eq:problem} \\
	J(A^k) & \triangleq \sum_{p\in {\cP^k}} \sum_{r\in \cR^k} a_{pr}^k \, \dotp{w_p(A^k)}{x_r^k} \\
	& = \dotp{A^k}{W(A^k)^\top X^k}, \label{eq:objective}
\end{align}
which provides a partial assignment of region descriptors $X^k$ to parts $W(A^k)$ in group $k$, such that the matching scores of matrix $M(A^k)$ are closely approximated by binary matrix $A^k$.

\subsection{Image grouping}
\label{sec:grouping}
The formulation above-given  refers to two problems: (i) grouping the images of the training set, and (ii) learning a set of discriminative parts per group. We follow a sequential approach by first grouping and then learning the parts, for each group independently. The latter is similar to the supervised setting of~\cite{Sicre15A,Sicre16A}, where classes are replaced by groups, and maintains the same complexity. We discuss grouping here and part learning in section~\ref{sec:parts}.

\head{Grouping by global similarity.}
Image grouping helps limiting the subsequent part learning into smaller training sets, but also specifying the objective of part learning, such that parts are discriminative according to requirement (iii) given in section~\ref{sec:formulation}. In this sense, images in a group should share patterns that do not occur in other groups.

Without referring to regions for this task, we follow the very simple solution of clustering images by global visual similarity. 
In particular, we represent each image $I$ in the training set $\cI$ by a global descriptor $x_I$ obtained by the last convolutional or fully connected layer of the same DCNN used to represent regions, see Section \ref{sec:details}. 
We then cluster $\cI$ in $K$ clusters using $k$-means on global representations in order to obtain a set of $k$ centroids $\{c^k: k \in [K]\}$. Finally, the clusters are balanced to obtain a uniform partition of the $N$ images of $\cI$ into $K$ groups of $N/K$ images each. The latter step is described below. The reason for balancing is twofold: (i) the cost of subsequent part learning is balanced, and (ii) each image receives the same weight, which is important since the number of parts per group is fixed.

\head{Greedy balancing.}
A simple form of balancing is to iterate over all groups, greedily assigning one image to a group at a time, until all images are assigned to a group. In particular, let $c^k$ be the centroid of cluster $k$ for $k \in [K]$. Also, let $\cU$ be the set of unassigned images, initially equal to $\cI$. For each $k \in [K]$, we choose the image $\arg\min_{I \in \cU} \norm{x_I - c^k}$ closest to $c^k$, assign it to group $\cI^k$ and remove it from $\cU$. We repeat this process until $\cU$ is empty.

\head{Iterative balancing.}
An alternative is to obtain a sequence of partitions $\cB_t$ of $\cI$, such that each partition $\cB_t$ is more balanced than the previous $\cB_{t-1}$, following~\cite{TaJA11}. Each partition $\cB_t = \{\cI_t^k: k \in [K]\}$ is defined by assigning each image $I \in \cI$ to the group $\arg\min_{k \in [K]} d_t(c^k, x_I)$, where $d_t(c, x)^2$ for $c, x \in \real^d$ is a penalized form of squared Euclidean distance, given by:
\begin{equation}
	d_t(c, x)^2 \triangleq \norm{c - x}^2 + b_t^k
	\label{eq:penalty-dist}
\end{equation}
 with $b_t^k$ a penalization term that is an increasing function of the cardinality $N_t^k = \card{\cI_t^k}$ of group $\cI_t^k$ at iteration $t$. In particular, this term is defined as $b_0^k = 1$ and
\begin{equation}
	b_t^k = b_{t-1}^k \left(\frac{N_t^k}{N/K}\right)^\alpha
	\label{eq:penalty-term}
\end{equation}
for $t > 0$ and $k \in [K]$. Then, the sequence $\cB_t$ converges to a uniform partition, i.e. $N_t^k \to N/K$ as $t \to \infty$ for $k \in [K]$, with parameter $\alpha$ controlling the speed of convergence. In practice, we get a partition $\cB = \{\cI^k: k \in [K]\}$ after 80 iterations with $\alpha=0.01$ \cite{TaJA11}.


\subsection{Learning parts per group}
\label{sec:parts}
Given a partition $\cB=\{\cI^k: k \in [K]\}$ of training images $\cI$, the optimization problem~(\ref{eq:problem}) is to be solved for each group $k$. The solution given in~\cite{Sicre15A} is iterative, alternating between optimizing region to part assignments $A^k$ and part models $W(A^k)$, keeping the other fixed. This is referred to as \emph{iterative soft-assignment} (\ISA).

On the other hand,~\cite{Sicre16A} substitutes (\ref{eq:lda-mat}) into (\ref{eq:objective}), resulting in a quadratic objective function with respect to $A^k$, with $W(A^k)$ eliminated. This opens the door to any algorithm for the \emph{quadratic assignment} problem. The \emph{Hungarian algorithm} (\HunA) is a particular case of non-iterative method examined in that work when matrix $M(A^k)$ is fixed.

While we use both \ISA and \HunA, we do not consider the quadratic assignment formulation of~\cite{Sicre16A} since we do not use any other iterative solution given in that work. We discuss the two approaches below.

\head{Iterative soft-assignment (\ISA).}
Starting from an initial matrix $A^k$, \ISA iteratively computes a part model matrix $W^k \gets W(A^k)$ for fixed $A^k$ by LDA using~(\ref{eq:lda-mat}), and optimizes cost function~(\ref{eq:objective}) to update $A^k$, keeping $W^k$ fixed.

The latter part is done in three steps. First, it applies soft-assignment to the matching matrix, $A^k \gets \sigma_\beta(M^k) = \sigma_\beta((W^k)^\top X^k)$
\begin{equation}
	\sigma_\beta(M^k) \triangleq \exp\{\beta(M^k - (\max_{r} M^k) \one_{R^k}^\top)\},
	\label{eq:soft-assign}
\end{equation}
where function $\exp$ is taken element-wise rather than matrix exponential, and $\max_r$ denotes row-wise maximum (over regions of an image). Function $\sigma_\beta$ is a form of soft-assignment that is scaled by parameter $\beta$ and only ensures that the row-wise $\ell^\infty$ norm is $1$. Second, $A^k$ is thresholded element-wise as $A_k \gets \tau(A^k)$ so that low values are set to zero. This is a means to achieve inequality constraint~(\ref{eq:cons-different}), as entire columns are gradually set to zero. Third, it iteratively normalizes rows and columns according to $\ell^1$ norm, until $A^k$ becomes bi-stochastic. This is the Sinkhorn algorithm, except that zero columns are left unnormalized.

The iteration given above optimizes a modified version of cost function~(\ref{eq:objective}) that includes a negative-entropy regularization term with coefficient $\frac{1}{\beta}$~\cite{Sicre16A} and satisfies constraints~(\ref{eq:cons-different}),(\ref{eq:cons-present}), but is not binary. The latter is achieved by repeating the entire process for increasing $\beta$. This yields a solution to problem~(\ref{eq:problem}) for $\beta \to \infty$, which is a form of \emph{deterministic annealing}.  

\head{Hungarian algorithm (\HunA).}
gives the exact solution of problem~(\ref{eq:problem}) assuming $W(A^k)$ (or $M(A^k)$) is fixed, which is a \emph{linear assignment} problem. In~\cite{Sicre16A}, \HunA has been used both as a standalone method and as part of an iterative algorithm, IPFP. \HunA is very fast compared to \ISA but with the limitation of assuming $M(A^k)$ fixed, it is expected to be inferior as a standalone solution.

The experimental results of~\cite{Sicre16A} show that \HunA competes with iterative IPFP and that both are inferior to \ISA, in terms of performance. However, we revisit this comparison with a new setup where \HunA actually competes with \ISA. This is an interesting finding, both for the efficiency of \HunA and the fact that it is not actually solving problem~(\ref{eq:problem}).

\subsection{Algorithm}
\label{sec:algorithm}
The entire algorithm of unsupervised part learning is summarized in Algorithm~\ref{alg:part-learn}. First, a global descriptor $x_I$ is computed for each image $I \in \cI$. These descriptors are then clustered into centroids $c^k$ for $k \in [K]$. Given both centroids and descriptors, we produce a uniform partition $\cB$ of $\cI$ into $K$ groups of $N/K$ images each, using either greedy (\textsc{greedy}) or iterative (\textsc{iter}) balancing, see section~\ref{sec:grouping}.

Then, we iterate over each group $\cI^k \in \cB$, beginning by computing region descriptors $X^k$. To initialize part models $W^k$ in a discriminative way, shown as \textsc{init-parts} in Algorithm~\ref{alg:part-learn}, we follow~\cite{Sicre15A}. In particular, descriptors in $X^k$ are clustered with $k$-means and, for each obtained centroid $c$ and its corresponding LDA model $w = \Sigma^{-1}(c-\mu)$, the max-pooled response $r_I^k(w) = \max_r w^\top X_I^k$ is computed for each image $I \in \cI$. These responses are summed over images in $\cI^k$ (resp. its complement in $\cI$) to yield a within-group response $r_+^k(w)$  (resp. between-group response $r_-^k(w)$). The $P$ models maximizing the within-group to between-group response ratio $r_+^k(w) / r_-^k(w)$ are chosen, represented in $d \times P$ matrix $W^k$. 

The remaining algorithm is independent per group $\cI^k$. Given initialized parts $W^k$, the matching matrix $M^k = (W^k)^\top X^k$ is computed and soft-assigned into $A^k$. \ISA or \HunA is applied on this $A^k$ and converts it to binary, solving problem~(\ref{eq:problem}). Algorithm~\ref{alg:part-learn} includes \ISA as a function, where the first Sinkhorn step is only needed for consistency with \HunA. \HunA can operate with $M(A^k)$ directly as its input, but we rather use $A^k$ instead in algorithm~\ref{alg:part-learn}. Finally, part models $W^k$ are obtained as $W(A^k)$ by LDA~(\ref{eq:lda-mat}) and collected over all groups.

Although part learning is independent per group, we remind that parts are discriminative according to our third requirement, due to discriminative initialization and LDA classifiers.

In section~\ref{sec:exp}, we experiment with both options \ie \textsc{greedy} and \textsc{iter} for balanced grouping, as well as both options for part learning, \ie \ISA and \HunA. We also experiment with different number of groups $K$, while the number of regions $|\cR|$ and parts $P$ are fixed. Although the focus of this work is on unsupervised part learning, we additionally experiment on supervised learning with an improved experimental setup, which is comparable to previous work~\cite{Sicre15A,Sicre16A}. Apart from classification, we additionally consider image retrieval as an end task.


\SetAlFnt{\small}
\begin{algorithm}[t]
\small
\DontPrintSemicolon
\SetEndCharOfAlgoLine{}
\SetFuncSty{textsc}
\newcommand{\commentsty}[1]{{\color{DarkGreen}#1}}
\SetCommentSty{commentsty}
\SetKwComment{Comment}{$\triangleright$ }{}
\SetKwBlock{Block}{function}{}

\SetKwFunction{Group}{group}
\SetKwFunction{Greedy}{greedy}
\SetKwFunction{Hun}{hun}
\SetKwFunction{InitParts}{init-parts}
\SetKwFunction{Isa}{isa}
\SetKwFunction{Iter}{iter}
\SetKwFunction{LearnParts}{learn-parts}
\SetKwFunction{Means}{means}
\SetKwFunction{Sinkhorn}{Sinkhorn}

\Block({$W \gets \LearnParts(\cI)$})
{
	Compute global descriptors $X \in {\mathbb R}^{d \times N}$ \\
	$C \gets k$-$\Means(X, K)$ \Comment*[f]{$k$-means clustering} \\
	$\cB \gets \Greedy(C, X) \text{ or } \Iter(C, X)$ \Comment*[f]{grouping, section~\ref{sec:grouping}} \\
	\For{$\cI^k \in \cB$}
	{
		Compute region descriptors $X^k \in {\mathbb R}^{d \times R^k}$ \\
		$W^k \gets \InitParts(X^k)$ \Comment*[f]{initial part descriptors} \\
		$A^k \gets \sigma_\beta((W^k)^\top X^k)$ \Comment*[f]{soft-assign (\ref{eq:soft-assign})} \\
		$A^k \gets \Isa(A^k, X^k) \text{ or } \Hun(A^k)$ \Comment*[f]{hard-assign} \\
		$W^k \gets W(A^k)$ \Comment*[f]{LDA~(\ref{eq:lda-mat})} \\
	}
	$W \gets \{W^k: k \in [K]\}$ \Comment*[f]{learned part models}
}

\Block({$A \gets \Isa(A, X)$})
{
	$A \gets \Sinkhorn(A)$ \Comment*[f]{make $A$ bi-stochastic} \\
	\For{$\beta \in \{\beta_0, \dots, \beta_{\max}\}$}
	{
		\While{$A$ not converged}
		{
			$W \gets W(A)$ \Comment*[f]{LDA~(\ref{eq:lda-mat})} \\
			$A \gets \tau(\sigma_\beta((W)^\top X))$ \Comment*[f]{soft-assign (\ref{eq:soft-assign})} \\
			$A \gets \Sinkhorn(A)$ \Comment*[f]{make $A$ bi-stochastic} \\
		}
	}
}

\caption{Unsupervised part learning}
\label{alg:part-learn}
\end{algorithm}

\section{Experiments}
\label{sec:exp}
This section presents an experimental validation of the above-presented method, applied to image classification as well as image retrieval. We first introduce the datasets, then provide implementation details, and finally present the results we obtained.

\subsection{Datasets}

\head{Willow actions~\cite{delaitre10}}
classification dataset contains 911 images split into 7 classes of common human actions, namely \textit{interacting with a computer, photographing, playing music, riding cycle, riding horse, running, walking}.
There are at least 108 images per action, with around 60 images used as training and the rest as testing images.
The dataset also offers bounding boxes, which are not used as we want to detect the relevant parts of images automatically.

\head{MIT 67 scenes~\cite{Quattoni09}}
aims at classifying indoor scenes and is composed of 67 classes. 
These include stores (e.g. bakery, toy store), home (e.g. kitchen, bedroom), public spaces (e.g. library, subway), leisure (e.g. restaurant, concert hall), and work (e.g. hospital, TV studio).
Each category has around 80 images for training and 20 for testing, totalling 6700 images.

\head{Oxford 5k~\cite{PCISZ07} and Paris 6k~\cite{PCISZ08}}
retrieval datasets contain 5,063 and 6,392 images respectively and have 55 query images each. Query and positive images depict landmarks of the two cities and there are 11 landmarks in each dataset with 5 queries each. Negatives are images from the same two cities but not depicting the landmarks. Performance is evaluated by means of mean Average Precision (mAP). Hard positive images are labeled as \emph{junk} and not taken into account in mAP computation.

\subsection{Implementation details}
\label{sec:details}

\head{Image regions}.
A set of proposed regions are obtained using Selective Search \cite{vandeSande11}, as in \cite{Sicre16A}. The total number of regions per image is fixed to $|\cR|=1,000$. If less than $|\cR|$ regions are available, we add random regions to reach $|\cR|$.

\head{Region descriptors}.
We use a number of DCNN image descriptors, choosing for each task the network giving state-of-the-art performance on the given datasets in the literature. Our motivation is indeed to show that our part-based model can improve on these very well performing networks.

On the Willow dataset, the last convolutional layer of the very deep VD19 network~\cite{Simonyan14c} is used for global representation and representation of regions.
We note that for this network, images are resized to 768 pixels maximum dimension and average pooling is performed to obtain a 512-dimensional description.

On MIT 67 scenes, the seventh fully connected layer of the very deep VD16 network trained on Places 205~\cite{zhou16} is used for region and global image description, giving a 4096-dimensional vector. PCA is applied at the encoding stage to reduce the description from 4096 to 512 dimensions.

Finally, for image retrieval, ResNet101~\cite{he16} fine-tuned on Landmarks dataset~\cite{gordo16} is used. The network includes max-pooling (MAC~\cite{Tolias16}), PCA, and normalization, and outputs a 2048-dimensional descriptor.

\head{Learning parts}.
We follow the general learning and classification pipeline of~\cite{Sicre15A}, replacing classes by computed groups in the case of unsupervised part learning.
Specifically, during part learning, $\card{\cR}=1,000$ regions are extracted from each image to learn $P = 100$ parts per group, both for classification and retrieval, while the number of groups $K$ is a varying parameter. We note that the parameters of the \ISA method to learn parts are identical to the ones used in \cite{Sicre15A}.


\head{Encoding}.
Once the parts are learned, the encoding stage aims at collecting part responses for a given image to build an image descriptor. 
In particular, for a given image, a set of $\card{\cR}=1,000$ regions and their descriptors are extracted as described previously. For each region descriptor $x$, the score $\dotp{w_p(A)}{x}$ of every part classifier $p$ of every group $k \in [K]$ is computed. We present here different encodings. 

The \emph{bag-of-parts} (BOP) is a $2PK$-dimensional descriptor, built by concatenating the average and maximum score over all image regions per part. A second option is to add the maximum score over each quarter image per part to the BOP. This $6PK$-dimensional descriptor is referred to as \emph{spatial bag-of-parts}~(sBOP).
As an alternative, each part is described by the descriptor of the region giving the maximum classification score over all image regions. These region DCNN descriptors are then concatenated, optionally being reduced by PCA before concatenating. This descriptor is referred to as \emph{PCAed CNN-on-parts} (pCOP) and has $d'PK$ dimensions, with $d'$ the dimension of the (reduced) DCNN descriptor.
In this work, we also propose to weigh the DCNN descriptor of the maximum scoring regions by their part classifier score, referred to as \emph{weighted} pCOP (wpCOP).
This encoding allows to combine information from both BOP and COP.
All descriptors and encoded representations are $\ell^2$-normalized.

\head{Classification pipeline}.
Parts are learned on the training images. Training and test images are then described by the same encoding method. Finally, a linear SVM is learned on the training set, and applied to classify test images.

\head{Retrieval pipeline}.
Parts are learned on images of the database. Database and query images are then described by the same encoding method. Finally, for each query, the database images are ranked by dot product similarity (all descriptors being $\ell^2$-normalized).

\subsection{Results}

This section presents an extensive study of the performance of part learning applied to both classification and retrieval. Although the focus and contribution of this work is unsupervised part learning, we also experiment on supervised part learning in the case of classification. This enables comparison of our improved pipeline (including different DCNN and encodings used) to previous work, providing new findings on the relative performance of algorithms like \ISA and \HunA.

\newcommand\multi[2]{\multicolumn{1}{|c|}{\multirow{#1}{*}{#2}}}
\begin{table}[tb]
    \small
	\centering
	\caption{Supervised part learning for classification, using different algorithms and encodings, compared to baseline global descriptors. S-\ISA: supervised \ISA; S-\HunA: supervised \HunA. }
	\label{tab:sup}
	\begin{tabular}{|l|c|cc|} \hline
	\multi{2}{Method}   &  Willow    &  \multicolumn{2}{c|}{MIT 67} \\ \cline{2-4}
	 & mAP & mAP & Acc.  \\ \hline\hline
Global & 88.5 & 83.6 & 78.5  \\ \hline
S-\ISA  BOP &  89.2 & 86.6 & 81.6 \\
S-\ISA  sBOP &  90.1 & 86.7 & 82.5 \\
S-\ISA  pCOP &  91.7 & 86.5 & 82.4 \\
S-\ISA  wpCOP &  \textbf{92.4} & \textbf{88.3} & \textbf{82.8} \\ \hline
S-\HunA BOP &  88.1 & 86.9 & 82.3 \\
S-\HunA sBOP  &  87.6 & 87.6 & 83.1 \\
S-\HunA pCOP & 91.1  & 86.2 & 81.9  \\
S-\HunA wpCOP &  \textbf{91.6}  & \textbf{88.8} & \textbf{83.7} \\ \hline
	\end{tabular}
\end{table}	

\head{Supervised part learning for classification.}
Global image descriptors are compared with two part learning methods, \ie \ISA and \HunA, using various encodings on two classification datasets. The process is exactly as shown in Algorithm~\ref{alg:part-learn} but using given classes instead of computed groups, similarly to previous work. Results are given in Table~\ref{tab:sup}.
We observe that part learning outperform global image representations on both datasets, with a larger gain on MIT 67.
Interestingly, \HunA outperforms \ISA on MIT 67, despite being inferior on Willow.
This is important since \HunA is not supposed to solve problem~(\ref{eq:problem}) exactly, but rather the special case of optimizing the part-to-region assignment when the part models are fixed.
Also note that \HunA is up to 100 times faster than \ISA, therefore favored in some of the experiments.
Furthermore, we show that the proposed wpCOP is the best performing encoding in all experiments.

\begin{table}[tb]
	\small
	\centering
	\caption{Unsupervised part learning for classification, using different matching algorithms and different encodings on Willow actions. Grouping performed with iterative balancing. Results given in terms of mAP. Performances  better than the baseline global descriptors are in bold.}
	\label{tab:will}
	\begin{tabular}{|c|l|cc|} \hline
	$K$ & Encoding & \ISA   & \HunA \\ \hline\hline
\multirow{2}{*}{5}  & sBOP & 77.5 & 75.4\\
                    & wpCOP & 88.4 & 86.8\\ \hline
\multirow{2}{*}{10} & sBOP & 81.9 & 77.5\\
                    & wpCOP & \textbf{89.5} & 88.4\\ \hline
\multirow{2}{*}{20} & sBOP & 85.9 & 81.8\\
                    & wpCOP & \textbf{90.4} & \textbf{89.3}\\ \hline
\multirow{2}{*}{40} & sBOP & 85.2 & 83.3\\
                    & wpCOP & \textbf{90.3} & \textbf{89.6}\\ \hline
\multirow{2}{*}{80} & sBOP & 85.3 & 84.7\\
                    & wpCOP & \textbf{88.8} & \textbf{89.1}\\ \hline

	\end{tabular}
\end{table}

\begin{table}[tb]
	\small
	\centering
	\caption{Unsupervised part learning for classification, using different matching algorithms and different encodings on MIT 67 Scenes. Grouping are performed with greedy and iterative balancing. Best scores are in bold}
	\label{tab:unsup}
	\begin{tabular}{|c|l|cc|cc|} \hline
		\multirow{2}{*}{$K$} & \multicolumn{1}{|c|}{\multirow{2}{*}{Method}} & \multicolumn{2}{|c|}{Greedy} & \multicolumn{2}{|c|}{Iterative} \\ \cline{3-6}
		                &             & mAP  & Acc. & mAP  & Acc. \\ \hline\hline
		\multi{2}{100}  & \ISA sBOP   & 86.2 & 81.0 & 85.6 & 81.2 \\
		                & \ISA wpCOP  & 87.8 & 82.4 & 87.6 & 82.3 \\ \hline
		\multi{2}{ 67}  & \ISA sBOP   & 85.9 & 81.5 & 85.5 & 80.2 \\
		                & \ISA wpCOP  & 87.8 & 82.2 & 87.5 & 81.7 \\ \hline
		\multi{2}{ 50}  & \ISA sBOP   & 85.0 & 80.3 & 85.6 & 80.1 \\
		                & \ISA wpCOP  & 86.8 & 81.9 & 87.7 & 81.3 \\ \hline\hline

		\multi{2}{100}  & \HunA sBOP  & 87.1 & 83.1 & 87.3 & \textbf{83.7} \\
		                & \HunA wpCOP & 88.6 & 83.2 & \textbf{88.8} & 83.4 \\ \hline
		\multi{2}{ 67}  & \HunA sBOP  & 86.6 & 82.3 & 86.8 & 83.5 \\
		                & \HunA wpCOP & \textbf{88.7} & \textbf{83.6} & \textbf{88.8} & 83.3 \\ \hline
		\multi{2}{ 50}  & \HunA sBOP  & 86.1 & 81.8 & 86.3 & 82.1 \\
		                & \HunA wpCOP & 88.1 & 82.9 & 87.6 & 82.0 \\ \hline

	\end{tabular}
\end{table}

\begin{table}[t]
	\small
	\centering
	\caption{Unsupervised part learning for classification on MIT 67 Scenes using VD19 for initialization. Grouping performed with iterative balancing.}
	\label{tab:init}
	\begin{tabular}{|l|l|c|c|} \hline
		$K$  & Method & mAP & Acc. \\ \hline\hline
		\multi{2}{100} & \HunA sBOP  & 85.9 & 81.6   \\
		               & \HunA wpCOP & 87.6 & 83.5\\ \hline
		\multi{2}{50}  & \HunA sBOP  & 85.1 & 80.6 \\
		               & \HunA wpCOP & 87.1 & 83.2 \\ \hline
	\end{tabular}
\end{table}

\head{Unsupervised part learning for classification.}
The proposed unsupervised part learning strategy is then evaluated on Willow in Table \ref{tab:will}, and on MIT 67 in Table \ref{tab:unsup}.
For these experiments, we retain two encodings: wpCOP for its higher performance (as shown in the previous experiment), and sBOP for its lower dimensionality. Each of the two encodings is combined with both algorithms \ISA and \HunA.
As for the case of supervised learning, we observe that none of the two algorithms really outperforms the other: \HunA outperforms \ISA on MIT 67 but is inferior on Willow.

Various numbers of groups are evaluated: $K \in \{5,10,20,40,80\}$ on Willow
and $K \in \{50, 67, 100\}$ on MIT~67.
We observe that 20 and 40 groups offer the best performance on Willow using wpCOP encoding.
Similarly, we observe on MIT 67 that overall $K=100$ outperforms $K=67$, which outperforms $K=50$.
Although not shown in the Tables, $K=200$ gives similar accuracy and slightly lower mAP on MIT 67 compared to $K=100$.

We also observe that unsupervised parts encoded with sBOP do not outperform the global representation on Willow. 
However, sBOP on MIT 67 and wpCOP on both datasets offer significant improvements.
Even though performance is not as high as supervised part learning, which is expected, we observe that unsupervised parts are about 2\% mAP below supervised parts on Willow and only 0.3\% accuracy below on MIT 67.

Table \ref{tab:unsup} also studies the two grouping methods, \ie iterative balancing \vs greedy balancing. None is offering a significant gain over the other.
We repeat the grouping computation a number of times to check the influence of the randomized $k$-means initialization. The largest difference observed over three runs on Willow is 0.6\% mAP for sBOP and 0.4\% mAP for wpCOP.

The impact of the initial grouping is further investigated, see Table \ref{tab:init}.
Here the initial grouping is performed using a different global description, \ie the output of the convolutional layers of the very deep VD19 Network.
The performance varies slightly, \ie mAP is overall 1\% lower and accuracy is stable with lower performances on sBOP but higher on wpCOP.

\begin{figure*}[tb]
\centering
\includegraphics[height = 2.8cm]{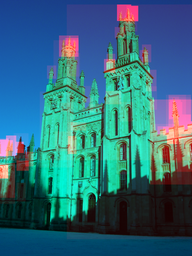} \ 
\includegraphics[height = 2.8cm]{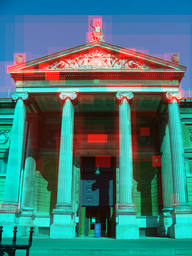} \ 
\includegraphics[height = 2.8cm]{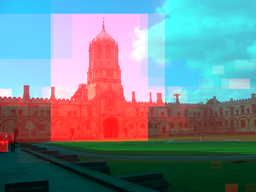} \ 
\includegraphics[height = 2.8cm]{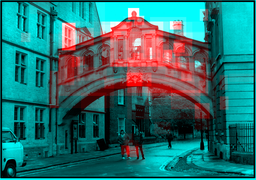} \
\includegraphics[height = 2.8cm]{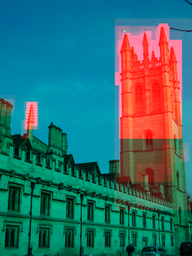} \ 
\includegraphics[height = 2.8cm]{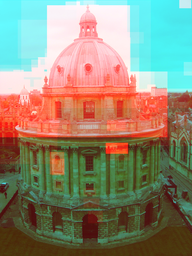} \\ \vspace{0.2cm}
\includegraphics[height = 2.8cm]{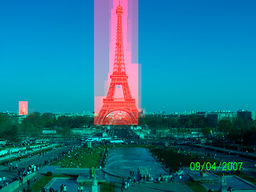} \ 
\includegraphics[height = 2.8cm]{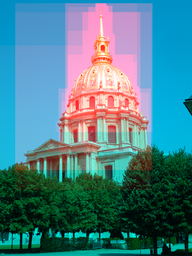} \ 
\includegraphics[height = 2.8cm]{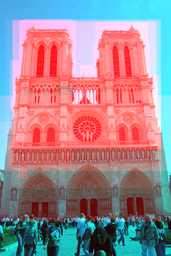} \ 
\includegraphics[height = 2.8cm]{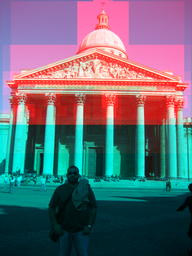} \ 
\includegraphics[height = 2.8cm]{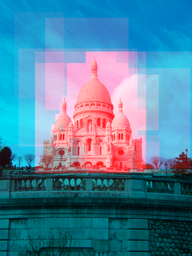} \ 
\includegraphics[height = 2.8cm]{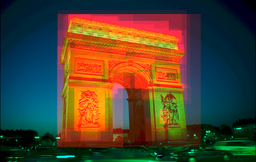} \ 
\small
\caption{The 200 top scoring parts are visualized on query images of Oxford5k (first row) and Paris6k (second row).}
\normalsize
\label{fig:ret}
\end{figure*}

\begin{table}[t]
	\small
	\centering
	\caption{Unsupervised part learning and mAP measurements on Oxford5k and Paris6k retrieval datasets. Grouping is performed with iterative balancing and part learning with \HunA.}
	\label{tab:ret}
	\begin{tabular}{|c|l|c|cc|} \hline
$K$             & Method & $d$ & Oxford5k & Paris6k \\ \hline\hline

                & \multi{5}{Global~\cite{gordo16}} & ori. & 83.2 & 92.4\\
                &                  & 512  & 78.9 & 92.5\\
                &                  & 256  & 76.2 & 90.9\\
                &                  & 128  & 73.0 & 89.0 \\
                &                  & 64   & 67.1 & 82.5\\  \hline \hline

\multi{5}{50}   & \multi{5}{\HunA sBOP}  & ori. & 77.2 & 91.4 \\
                &                  & 512  & 77.4 & 91.4 \\
                &                  & 256  & 77.1 & 91.5 \\
                &                  & 128  & 75.0 & 91.7 \\
                &                  & 64   & 71.6 & 91.5 \\ \hline
\multi{5}{50}   & \multi{5}{\HunA wpCOP} & ori. & 83.1 & 94.8 \\
                &                  & 512  & 84.3 & 94.6 \\
                &                  & 256  & 84.3 & 94.5 \\
                &                  & 128  & 81.7 & 94.3 \\
                &                  & 64   & 71.0 & 95.3 \\ \hline\hline

\multi{5}{100}  & \multi{5}{\HunA sBOP}  & ori. & 79.1 & 90.5 \\
                &                  & 512  & 79.0 & 90.5 \\
                &                  & 256  & 78.6 & 90.6 \\
                &                  & 128  & 77.5 & 90.7 \\
                &                  & 64   & 73.9 & 91.3 \\ \hline
\multi{5}{100}  & \multi{5}{\HunA wpCOP} & ori. & 83.5 & 94.5 \\
                &                  & 512  & 84.4 & 94.2 \\
                &                  & 256  & 84.1 & 94.0 \\
                &                  & 128  & 81.6 & 93.8 \\
                &                  & 64   & 69.6 & 94.0 \\ \hline
	\end{tabular}
\end{table}

\head{Unsupervised part learning for retrieval.}
Now, since our proposed part learning approach is unsupervised, it allows to learn parts without any labels.
Therefore we can apply this method on various tasks, where no annotated data is available, such as image retrieval. 
Table \ref{tab:ret} shows the performance of unsupervised part learning on the two image retrieval datasets, \ie Oxford5k and Paris6k.
Unsupervised part learning methods are compared against global image representations, as well as reduced representations. Having highly reduced representation is important in image retrieval for efficient search in large databases.

Although sBOP encoding is inferior to the global representation in the original descriptor dimensionality, wpCOP offers an improvement, which is larger when reducing the dimensionality.
We further note that sBOP outperforms global representation at low dimensionality and even outperforms wpCOP in the extreme case of 64 dimensions on oxford5k.
The gain in performance for unsupervised parts is observed on both datasets.
Interestingly, $K=100$ groups performs better than $K=50$ on Oxford5k, as for classification on MIT 67, but $K=50$ performs better on Paris6k.

Additionally, qualitative results are shown in Figure \ref{fig:ret}, where the 200 highest scoring parts are visualized on several query images of Oxford5k and Paris6k. Visualization of all query images appears in the supplementary material.

\begin{table}[tb]
	\small
	\centering
	\caption{Summary of our best results for supervised and unsupervised part learning on both classification and retrieval tasks. S-\ISA: supervised \ISA; S-\HunA: supervised \HunA. Unsupervised part learning is performed with $K=100$.}
	\label{tab:recap}
	\begin{tabular}{|l|c|cc|c|c|} \hline
		\multi{2}{Method} & \multi{2}{$d$} &  \multicolumn{2}{c|}{MIT 67} & Oxf5k & Paris6k\\ \cline{3-6}
	           &     & mAP  & Acc. & \multicolumn{2}{|c|}{mAP}  \\ \hline\hline

        Global &     & 83.6 & 78.5 & 83.2 & 92.4 \\
        Global & 256 & ---  & ---  & 76.2 & 90.9 \\ \hline
S-\HunA sBOP   &     & 87.6 & 83.1 &  --- &  --- \\
S-\HunA wpCOP  &     & 88.8 & 83.7 &  --- &  --- \\ \hline
  \HunA sBOP   &     & 87.3 & 83.7 & 79.1 & 90.5 \\
  \HunA wpCOP  &     & 88.8 & 83.4 & 83.5 & 94.5 \\ \hline
  \HunA sBOP   & 256 & --- & --- & 78.6 & 90.6 \\
  \HunA wpCOP  & 256 & --- & --- & 84.1 & 94.0 \\ \hline

	\end{tabular}
\end{table}

\head{Summary.}
Finally, Table \ref{tab:recap} summarizes our best results on MIT~67, Oxford5k, and Paris6k for the learning configuration using $K=100$ groups with grouping by iterative balancing and part learning by \HunA. It is clear that \HunA and \ISA are two comparable part learning approaches with \HunA being faster to compute. It is remarkable that the same part learning approaches are competitive both in a supervised and an unsupervised setup. Our proposed wpCOP encoding outperforms all alternatives. There is a clear gain in using part-based models in classification, even in unsupervised fashion, compared to global representations. In retrieval, gain is also obtained in low dimensionality.

\section{Conclusions}
\label{sec:ccl}

This paper introduces a novel framework for the unsupervised learning of part-based models.
The key idea is to generate groups of similar images, through the use of a clustering algorithm, and learn part models that are discriminative w.r.t. the different groups. 
Our intuition is that our part learning method is capable of capturing the data distribution for a novel task without requiring any labels for this task.
We demonstrate that our part-based models, when used to encode images, improve the performance of image classifiers compared to a global encoding of images. More importantly, these models open the door to new applications for which no class labels are available, \eg instance retrieval. Our approach is experimentally validated on two classification and two retrieval datasets, consistently improving the performance of the state-of-the art DCNNs.

{\small
\bibliographystyle{ieee}
\bibliography{biblio}

\begin{thebibliography}{10}\itemsep=-1pt

\bibitem{AKTS10}
Y.~Avrithis, Y.~Kalantidis, G.~Tolias, and E.~Spyrou.
\newblock Retrieving landmark and non-landmark images from community photo
  collections.
\newblock In {\em ACM Multimedia}, pages 153--162. ACM, 2010.

\bibitem{ChMa10}
O.~Chum and J.~Matas.
\newblock Large-scale discovery of spatially related images.
\newblock {\em Pattern Analysis and Machine Intelligence, IEEE Transactions
  on}, 32(2):371--377, Feb 2010.

\bibitem{delaitre10}
V.~Delaitre, I.~Laptev, and J.~Sivic.
\newblock Recognizing human actions in still images: a study of bag-of-features
  and part-based representations.
\newblock In {\em Proceedings of the British Machine Vision Conference},
  volume~2, 2010.

\bibitem{doersch13}
C.~Doersch, A.~Gupta, and A.~A. Efros.
\newblock Mid-level visual element discovery as discriminative mode seeking.
\newblock In {\em Advances in Neural Information Processing Systems}, pages
  494--502, 2013.

\bibitem{felzenszwalb10}
P.~F. Felzenszwalb, R.~B. Girshick, D.~McAllester, and D.~Ramanan.
\newblock Object detection with discriminatively trained part-based models.
\newblock {\em {\sc IEEE} Transactions on Pattern Analysis and Machine
  Intelligence}, 32(9):1627--1645, 2010.

\bibitem{gordo16}
A.~Gordo, J.~Almazan, J.~Revaud, and D.~Larlus.
\newblock End-to-end learning of deep visual representations for image
  retrieval.
\newblock {\em arXiv preprint arXiv:1610.07940}, 2016.

\bibitem{he16}
K.~He, X.~Zhang, S.~Ren, and J.~Sun.
\newblock Deep residual learning for image recognition.
\newblock {\em CVPR}, 2016.

\bibitem{HGO+10}
K.~Heath, N.~Gelfand, M.~Ovsjanikov, M.~Aanjaneya, and L.~J. Guibas.
\newblock Image webs: Computing and exploiting connectivity in image
  collections.
\newblock In {\em IEEE Conference on Computer Vision and Pattern Recognition},
  2010.

\bibitem{Juneja13}
M.~Juneja, A.~Vedaldi, C.~V. Jawahar, and A.~Zisserman.
\newblock Blocks that shout: Distinctive parts for scene classification.
\newblock In {\em Proceedings of the IEEE Conference on Computer Vision and
  Pattern Recognition}, 2013.

\bibitem{Kulkarni16}
P.~Kulkarni, F.~Jurie, J.~Zepeda, P.~P\'erez, and L.~Chevallier.
\newblock {SPLeaP: Soft Pooling of Learned Parts for Image Classification}.
\newblock In {\em ECCV}, 2016.

\bibitem{Leibe2008}
B.~Leibe, A.~Leonardis, and B.~Schiele.
\newblock Robust object detection with interleaved categorization and
  segmentation.
\newblock {\em Int. J. Comput. Vision}, 77(1-3):259--289, May 2008.

\bibitem{Mettes15}
P.~Mettes, J.~C. van Gemert, and C.~G.~M. Snoek.
\newblock No spare parts: Sharing part detectors for image categorization.
\newblock {\em CoRR}, abs/1510.04908, 2015.

\bibitem{parizi2014automatic}
S.~N. Parizi, A.~Vedaldi, A.~Zisserman, and P.~Felzenszwalb.
\newblock Automatic discovery and optimization of parts for image
  classification.
\newblock In {\em International Conference on Learning Representations}, 5
  2015.

\bibitem{PCISZ07}
J.~Philbin, O.~Chum, M.~Isard, J.~Sivic, and A.~Zisserman.
\newblock Object retrieval with large vocabularies and fast spatial matching.
\newblock In {\em Proceedings of the IEEE Conference on Computer Vision and
  Pattern Recognition}, June 2007.

\bibitem{PCISZ08}
J.~Philbin, O.~Chum, M.~Isard, J.~Sivic, and A.~Zisserman.
\newblock Lost in quantization: Improving particular object retrieval in large
  scale image databases.
\newblock In {\em Proceedings of the IEEE Conference on Computer Vision and
  Pattern Recognition}, June 2008.

\bibitem{QuLV08}
T.~Quack, B.~Leibe, and L.~Van~Gool.
\newblock World-scale mining of objects and events from community photo
  collections.
\newblock In {\em Conference on Image and Video retrieval}, pages 47--56, 2008.

\bibitem{Quattoni09}
A.~Quattoni and A.~Torralba.
\newblock Recognizing indoor scenes.
\newblock In {\em Proceedings of the IEEE Conference on Computer Vision and
  Pattern Recognition}, 2009.

\bibitem{RaTC16}
F.~Radenovic, G.~Tolias, and O.~Chum.
\newblock {CNN} image retrieval learns from {BoW}: Unsupervised fine-tuning
  with hard examples.
\newblock In {\em European Conference on Computer Vision}, 2016.

\bibitem{Sicre15B}
R.~Sicre and H.~J{\'e}gou.
\newblock Memory vectors for particular object retrieval with multiple queries.
\newblock In {\em Proceedings of the 5th ACM on International Conference on
  Multimedia Retrieval}, pages 479--482. ACM, 2015.

\bibitem{Sicre14parts}
R.~Sicre and F.~Jurie.
\newblock Discovering and aligning discriminative mid-level features for image
  classification.
\newblock In {\em International Conference on Pattern Recognition}, pages
  1975--1980. IEEE, 2014.

\bibitem{Sicre15A}
R.~Sicre and F.~Jurie.
\newblock Discriminative part model for visual recognition.
\newblock {\em Computer Vision and Image Understanding}, 141:28 -- 37, 2015.

\bibitem{Sicre16A}
R.~Sicre, J.~Rabin, Y.~Avrithis, T.~Furon, and F.~Jurie.
\newblock Automatic discovery of discriminative parts as a quadratic assignment
  problem.
\newblock {\em arXiv preprint arXiv:1611.04413}, 2016.

\bibitem{Simonyan14c}
K.~Simonyan and A.~Zisserman.
\newblock Very deep convolutional networks for large-scale image recognition.
\newblock {\em CoRR}, abs/1409.1556, 2014.

\bibitem{singh12}
S.~Singh, A.~Gupta, and A.~A. Efros.
\newblock Unsupervised discovery of mid-level discriminative patches.
\newblock In {\em Proceedings of the European Conference on Computer Vision},
  pages 73--86. Springer, 2012.

\bibitem{TaJA11}
R.~Tavenard, H.~Jegou, and L.~Amsaleg.
\newblock Balancing clusters to reduce response time variability in large scale
  image search.
\newblock In {\em Content-Based Multimedia Indexing}, 2011.

\bibitem{Tolias16}
G.~Tolias, R.~Sicre, and H.~J{\'{e}}gou.
\newblock Particular object retrieval with integral max-pooling of {CNN}
  activations.
\newblock 2016.

\bibitem{TuLo09}
P.~Turcot and D.~Lowe.
\newblock Better matching with fewer features: the selection of useful features
  in large database recognition problems.
\newblock In {\em International Conference on Computer Vision}, 2009.

\bibitem{Ullman2001gh}
S.~Ullman, E.~Sali, and M.~Vidal-Naquet.
\newblock {A Fragment-Based Approach to Object Representation and
  Classification}.
\newblock In {\em Visual Form 2001}, pages 85--100. Springer Berlin Heidelberg,
  Berlin, Heidelberg, 2001.

\bibitem{vandeSande11}
K.~E.~A. van~de Sande, J.~R.~R. Uijlings, T.~Gevers, and A.~W.~M. Smeulders.
\newblock Segmentation as selective search for object recognition.
\newblock In {\em IEEE International Conference on Computer Vision}, 2011.

\bibitem{weber10}
M.~Weber, M.~Welling, and P.~Perona.
\newblock Towards automatic discovery of object categories.
\newblock In {\em 2010 IEEE Conference on Computer Vision and Pattern
  Recognition}, 2010.

\bibitem{WeLe11}
T.~Weyand and B.~Leibe.
\newblock Discovering favorite views of popular places with iconoid shift.
\newblock In {\em International Conference on Computer Vision}, 2011.

\bibitem{zhou16}
B.~Zhou, A.~Khosla, A.~Lapedriza, A.~Torralba, and A.~Oliva.
\newblock Places: An image database for deep scene understanding.
\newblock {\em arXiv preprint arXiv:1610.02055}, 2016.

\end{thebibliography}
}



\end{document}